\documentclass[runningheads]{llncs}

\usepackage[mobile]{eccv}

\usepackage{eccvabbrv}

\usepackage{graphicx}
\usepackage{booktabs}
\usepackage{multirow}
\usepackage{tabularx}
\newcolumntype{C}{>{\centering\arraybackslash}X} 
\usepackage{array}

\usepackage[accsupp]{axessibility}  

\usepackage[pagebackref,breaklinks,colorlinks,citecolor=eccvblue]{hyperref}

\usepackage{orcidlink}

\begin{document}

\title{Exploring Pre-trained Text-to-Video Diffusion Models for Referring Video Object Segmentation} 

\titlerunning{Abbreviated paper title}

\author{Zixin Zhu\inst{1} \and
Xuelu Feng\inst{1} \and
Dongdong Chen\inst{2} \and
Junsong Yuan\inst{1} \and
Chunming Qiao\inst{1} \and Gang Hua\inst{3}\\
}

\authorrunning{F.~Author et al.}

\institute{{$^1$University at Buffalo} \quad  {$^2$Microsoft GenAI} \quad {$^3$Dolby Laboratories} \\
\email{\{zixinzhu,xuelufen,jsyuan,qiao\}@buffalo.edu, 
 {\{cddlyf,ganghua\}@gmail.com}}}

\maketitle

\begin{abstract}
  In this paper, we explore the visual representations produced from a pre-trained text-to-video (T2V) diffusion model for video understanding tasks. We hypothesize that the latent representation learned from a pretrained generative T2V model encapsulates rich semantics and coherent temporal correspondences, thereby naturally facilitating video understanding. Our hypothesis is validated through the classic referring video object segmentation (R-VOS) task. We introduce a novel framework, termed ``VD-IT'', tailored with dedicatedly designed components built upon a fixed pretrained T2V model. Specifically, VD-IT uses textual information as a conditional input, ensuring semantic consistency across time for precise temporal instance matching. It further incorporates image tokens as supplementary textual inputs, enriching the feature set to generate detailed and nuanced masks. Besides, instead of using the standard Gaussian noise, we propose to predict the video-specific noise with an extra noise prediction module, which can help preserve the feature fidelity and elevates segmentation quality. Through extensive experiments, we surprisingly observe that fixed generative T2V diffusion models, unlike commonly used video backbones (e.g., Video Swin Transformer) pretrained with discriminative image/video pre-tasks, exhibit better potential to maintain semantic alignment and temporal consistency. On existing standard benchmarks, our VD-IT achieves highly competitive results, surpassing many existing state-of-the-art methods. The code is available at \url{https://github.com/buxiangzhiren/VD-IT}.
  
\end{abstract}

\vspace{-0.5em}
\section{Introduction}
\label{sec:intro}
\vspace{-0.5em}
Recent studies on pre-trained text-to-image diffusion models~\cite{ramesh2022hierarchical,rombach2022high} have sparked growing interest in various image understanding tasks~\cite{wu2023diffumask,li2023open,xu2023open,pnvr2023ld,zhang2024tale}. These studies focus on analyzing and understanding the internal representations for individual images. Specifically, 
The benefits of pre-trained diffusion models, particularly their ability to provide open vocabulary knowledge and improve semantic differentiation between objects in the same image, are well investigated.

However, the potential of pre-trained text-to-video diffusion models~\cite{blattmann2023align,molad2023dreamix,zhou2022magicvideo,wang2023modelscope} in video understanding tasks~\cite{wang2023masked,wang2023look,wang2023omnitracker,wang2024omnivid} remains much less explored. Unlike image understanding, which primarily deals with single images, video comprehension necessitates the consideration of both spatial information within each frame and temporal information across all video frames, making it more challenging. Adhering to the principle of \emph{``what I cannot create, I do not understand''}, we hypothesize that a pre-trained text-to-video diffusion model capable of generating coherent, high-quality video sequences based on text prompts inherently possesses sufficient requisite knowledge to help video understanding tasks.

\begin{figure}[tb]
  \centering
  \begin{subfigure}[c]{0.59\linewidth}
     \includegraphics[width=\textwidth]{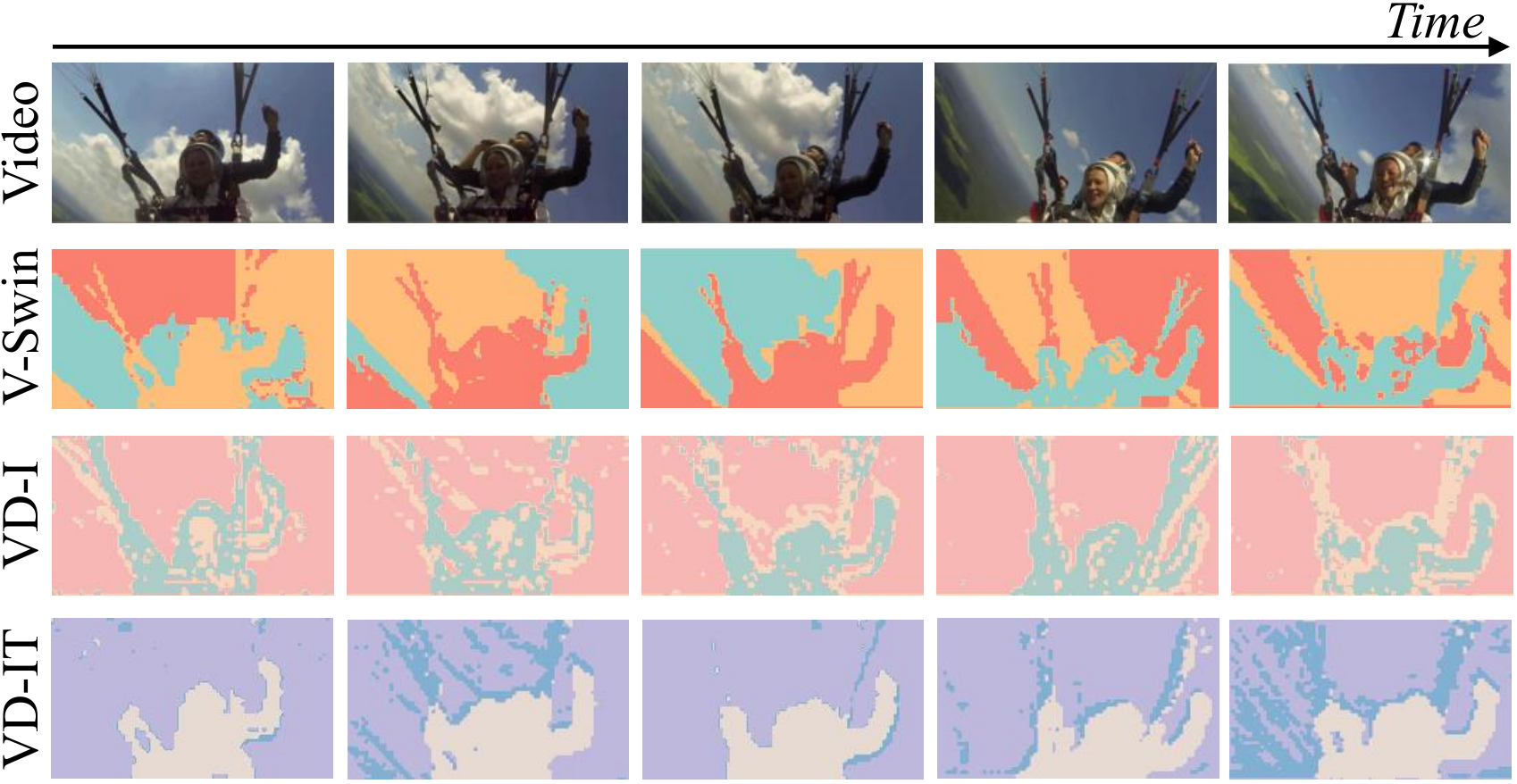}
  \end{subfigure}
  \hfill
  \begin{subfigure}[c]{0.34\linewidth}
     \includegraphics[width=\textwidth]{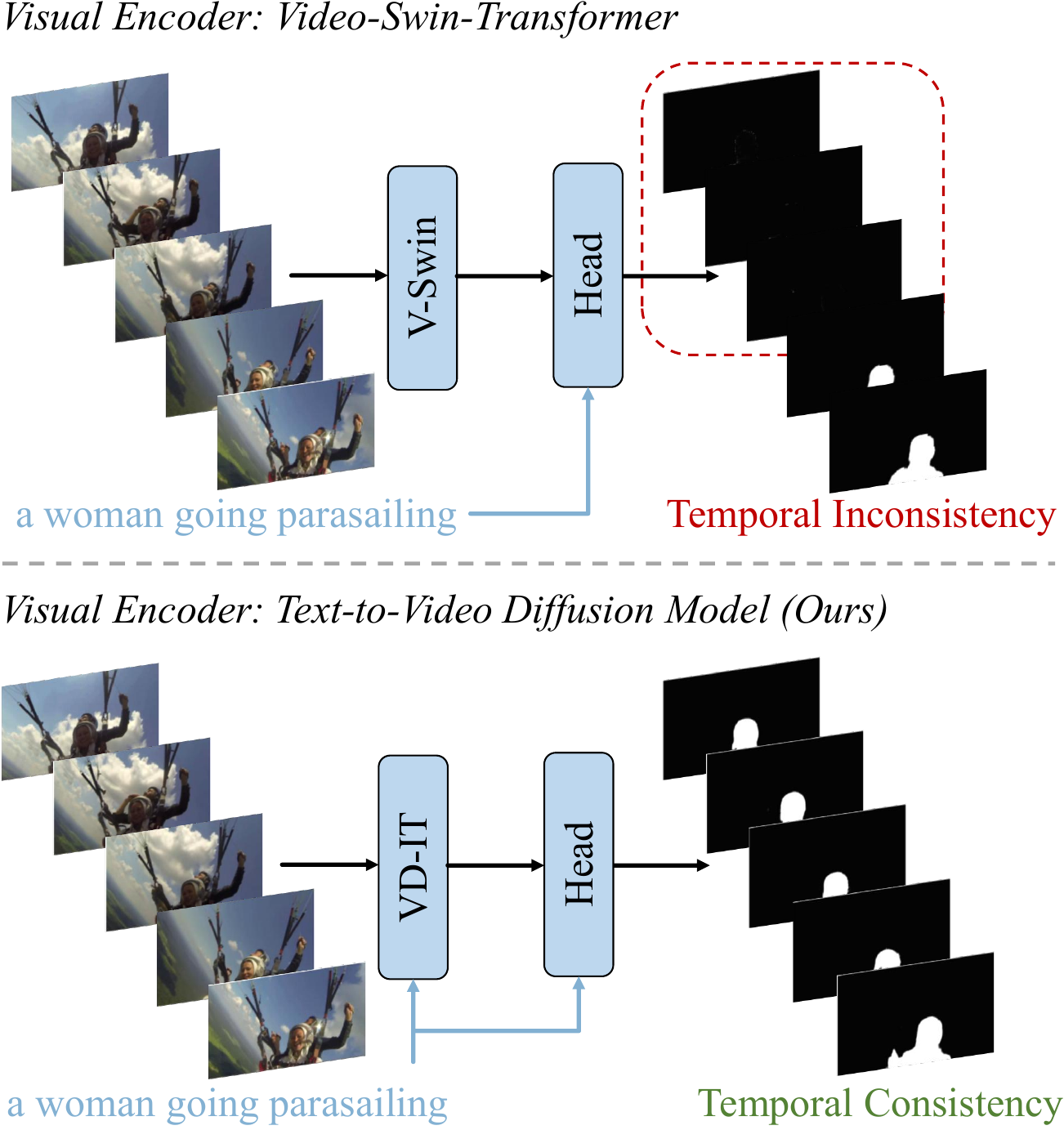}
  \end{subfigure}
  \vspace{-1em}
 \caption{Left: Analysis of learned features of existing  methods that  use discriminative backbone (Video Swin Transformer) and our methods (VD-I and VD-IT) that use fixed pretrained generative T2V model. Right: Temporal inconsistency in visual features will subsequently cause temporally inconsistent masks. }
  \label{fig:motivation}
  \vspace{-0.8cm} 
\end{figure}

We validate our hypothesis based on a popular text-to-video model ModelScopeT2V~\cite{wang2023modelscope} for the classic task of Referring Video Object Segmentation (R-VOS). R-VOS, guided by descriptive natural language prompts, aiming to delineate a specific object within a video sequence in a temporally consistent way. It has a variety of applications, such as video Editing and Video Retrieval. Over recent years, many R-VOS methods~\cite{mei2024slvp,miao2023spectrum,wu2022language,wu2023onlinerefer} have been proposed and achieved state-of-the-art performance. But most video backbone (e.g., Video Swin Transformer~\cite{liu2022video}) used in these methods are  pretrained with discriminative (e.g., classification) pre-tasks, and then need to be jointly finetuned with the text-driven mask decoder. Different from these methods, we explore directly exploiting the latent feature representation of a fixed generative text-to-video diffusion model, offering a fresh perspective in this domain.

In this paper, by analyzing the task characteristic of R-VOS,  we propose a new R-VOS framework termed ``VD-IT,'' built upon the open-sourced text-to-video diffusion model~\cite{wang2023modelscope}. VD-IT comprises two core components: video feature extraction and a mask decoder responsible for generating final object masks based on the provided text query and extracted video features. To ensure temporal consistency and enhance feature detail for high-quality mask generation, we incorporate two key designs in feature extraction from the text-to-video diffusion model: Text-Guided Image Projection and Video-specific Noise Prediction. Specifically, for Text-Guided Image Projection, we propose leveraging both referring text and visual tokens from each frame to guide the T2V model in producing the latent feature, rather than solely relying on visual tokens. This approach not only ensures visual feature consistency across time, aiding temporal instance matching but also enriches feature detail for better spatial differentiation. The latter, Noise Prediction, diverges from the standard Gaussian noise commonly applied in diffusion, predicting specific noise patterns during the diffusion forward process to further preserve the fidelity of extracted video features.

We evaluate the effectiveness of VD-IT on four standard R-VOS benchmarks: Ref-Youtube-VOS~\cite{seo2020urvos}, Ref-DAVIS17~\cite{khoreva2019video}, A2D-Sentences~\cite{gavrilyuk2018actor}, and JHMDB-Sentences \cite{gavrilyuk2018actor}. The quantitative and qualitative results demonstrate that our VD-IT achieves highly competitive performances and outperforms previous state-of-the-art methods. More interestingly, through in-depth analysis, we observe that the visual features learned in VD-IT exhibit significantly better temporal semantic consistency and spatial smoothness than existing methods~\cite{miao2023spectrum} that employ discriminatively finetuned video backbone (e.g., Video Swin Transformer). In Figure~\ref{fig:motivation}, we present the K-Means clustering results of the visual features coming from our methods (VD-IT and another variant VD-I that only uses visual tokens as guide the latent feature extraction) and one existing method \cite{miao2023spectrum} that uses fine-tuned Video Swin Transformer (V-Swin). It can be seen that the V-Swin feature changes significantly due to variations in lighting, affecting feature consistency across frames and complicating segmentation tasks. In contrast, VD-I and VD-IT maintain strong temporal consistency, with VD-IT further reducing internal noise through text guidance.

We believe that the superior temporal consistency observed in the T2V diffusion models can be attributed to the use of global text prompts as conditional inputs, which guides the generation of semantically consistent image frames throughout a video. This conditioning ensures that the semantic attributes across frames are aligned, enhancing temporal coherence. Additionally, the intrinsic denoising capability of the diffusion model bolsters its robustness against external disturbances like illumination changes and camera movement. By leveraging these mechanisms, the diffusion features exhibit enhanced temporal semantic consistency, proving advantageous for video understanding tasks. This idea is supported by analysis in following Section~\ref{method:IT} and Section~\ref{lightex}. It is crucial to clarify that our contribution involves more than merely substituting a visual backbone from V-Swin to a text-to-video diffusion model for enhanced performance. In fact, in the subsequent analysis section, we demonstrate that without our dedicated designs, simply utilizing the initial T2V diffusion model cannot achieve superior performances compared to existing methods. The main contributions of this work are:
\begin{itemize}
    \item To our knowledge, we are the first to explore the video prior (temporal consistency) acquired by pre-trained text-to-video diffusion models for video understanding task. Our findings reveal that, compared to conventional discriminatively fine-tuned video encoders, pretrained text-to-video diffusion models exhibit superior temporal consistency.
    \item We propose a new R-VOS framework VD-IT with several innovative designs, which significantly improves the extracted feature quality and boost the final performances. 
    \item Extensive experiments and analysis on Ref-Youtube-VOS ~\cite{seo2020urvos}, Ref-DAVIS17 \cite{khoreva2019video}, A2D-Sentences and JHMDB-Sentences~\cite{gavrilyuk2018actor} show that VD-IT can achieve super competitive results and outperform previous methods.
\end{itemize}

\section{Related Work}
\vspace{-0.5em}
\noindent\textbf{Diffusion Model.} 
In recent years, diffusion models have exhibited remarkable efficacy across various generative tasks, markedly enhancing generation quality, particularly in text-to-image \cite{rombach2022high,saharia2022palette,ramesh2022hierarchical,gu2022vector,wang2022semantic,fan2023frido,zhao2023uni} and text-to-video generation \cite{blattmann2023align,molad2023dreamix,zhou2022magicvideo,blattmann2023stable}. Moreover, there have been recent endeavors to design specialized diffusion models tailored for diverse image and video understanding tasks \cite{chen2023diffusion,tur2023exploring,zhao2023diffusionvmr,chen2023diffusiondet}. While these methods approach understanding tasks from a generative perspective, other works \cite{wu2023diffumask,li2023open,xu2023open,pnvr2023ld,zhang2024tale} directly leverage feature priors learned in pretrained text-to-image models for image understanding, showcasing the discriminative capabilities of latent feature representations learned in pretrained text-to-image models. However, to the best of our knowledge, our study marks the first exploration of temporal-consistency priors in pre-trained text-to-video diffusion models for video understanding tasks.

\noindent\textbf{Referring Video Object Segmentation.} 
In contrast to traditional semi-supervised video object segmentation methods~\cite{zhu2021separable,park2021learning}, relying on ground-truth mask annotations in the initial frame for target identification, R-VOS aims to delineate the object based on textual queries, demanding a nuanced understanding of cross-modal sources—vision and language. Previous R-VOS approaches~\cite{liang2021rethinking,miao2023spectrum,wu2023onlinerefer} leverage deep neural networks for vision-and-language interaction, enhancing visual features with linguistic information for pixel-level segmentation. For instance, \cite{liang2021rethinking} introduces a top-down R-VOS approach addressing limitations of bottom-up strategies, while SgMg \cite{miao2023spectrum} performs direct segmentation on encoded features, refining masks to combat drift. Diverging from discriminative feature representation learning in these methods , this paper studies a new perspective, i.e., directly leverage the feature representation learned in generative text-to-video model for video understanding. And the technique advancements in previous methods (e.g., better mask decoder design) are complementary to our method.

\vspace{-0.5em}
\section{Preliminary of Text-to-Video Diffusion Model}\label{preliminary}
\vspace{-0.5em}
In this section, we provide a brief overview of T2V diffusion models and describe how we extract features for video understanding tasks.

\noindent\textbf{Overview.} Given a text prompt, T2V diffusion models is designed to generate a coherent video based on the semantic meaning of the prompt. The overall architecture often consists of three parts: an encoder $\mathcal{E}$, a decoder $\mathcal{D}$ that is derived from VQGAN/VQVAE~\cite{esser2021taming}, translating image pixels into/back from latent spaces respectively, and a denoising U-Net $\mathcal{U}$ functioning in the latent space. The training procedure of T2V diffusion models is a forward-backward process. Specifically, the input video is first mapped to a latent vector. Then the latent vector is distorted by adding a small level of Gaussian noise. Meanwhile, the text prompt is encoded into a text embedding with a pre-trained text encoder~\cite{radford2021learning}. The denoising U-Net is optimised to predict the noise and recover the initial video signal, conditioned on the text embedding and the noised video. In the inference stage, the denoising U-Net takes video-shaped pure Gaussian noise and the text embedding of a user provided description as input, and progressively de-noises it to a realistic video through many iterations.

\noindent\textbf{Visual Representation Extraction.}\label{pre-vre} 
The denoising U-Net architecture is structured with downsampling, spatio-temporal, and upsampling blocks. It integrates cross-attention~\cite{vaswani2017attention} mechanisms that align text embeddings with U-Net features across spatial dimensions, alongside self-attention~\cite{vaswani2017attention} mechanisms that process these features through the temporal dimension for enhanced context comprehension. Thus the feature maps produced by each U-Net block encapsulate semantically aligned and temporally consistent attributes. To generate the latent features of the denoising U-Net for video understanding, it requires two inputs: the noise-added video and the text embedding. Following~\cite{xu2023open,zhang2024tale}, we introduce a single step of low-intensity noise to the video, aiming to preserve the original video signal as much as possible. However, our approach diverges from them~\cite{xu2023open,zhang2024tale} for image segmentation, where image projection serves as an implicit text embedding to circumvent the necessity for explicitly captioned image data. In our case, we employ both referring text and image projection, where referring text is to ensure temporal consistency across visual features while image projection is utilized as a supplementary condition to enhance the richness and detail of these features. Furthermore, in contrast to the referenced methods~\cite{xu2023open,zhang2024tale} that add pure Gaussian noise into the image, we leverage an extra noise prediction module that learns video-specific noise, which can help retain the fidelity in the generated video feature.

\vspace{-0.5em}
\section{Text-to-Video Diffusion Model for R-VOS}
\vspace{-0.5em}
In this paper, we use the referring video segmentation task as an example to explore the potential of learned representation of generative text-to-video models for video understanding tasks. Considering the pretrained text-to-video model ModelScopeT2V~\cite{wang2023modelscope} was open-sourced at the beginning of this project, we use it by default in all following experiments. Given a video clip $\mathcal{I}=\left\{I_t\right\}_{t=1}^T$ comprising $T$ frames, R-VOS task aims to generate $T$-frame binary segmentation masks for the referred object $\mathcal{S}=\left\{s_t\right\}_{t=1}^T, s_t \in \mathbb{R}^{H \times W}$, guided by a referring expression $\mathcal{E}=\left\{e_l\right\}_{l=1}^L$ consisting of $L$ words, in an temporally coherent way. 

\begin{figure}[tb]
  \centering
  \includegraphics[width=\textwidth]{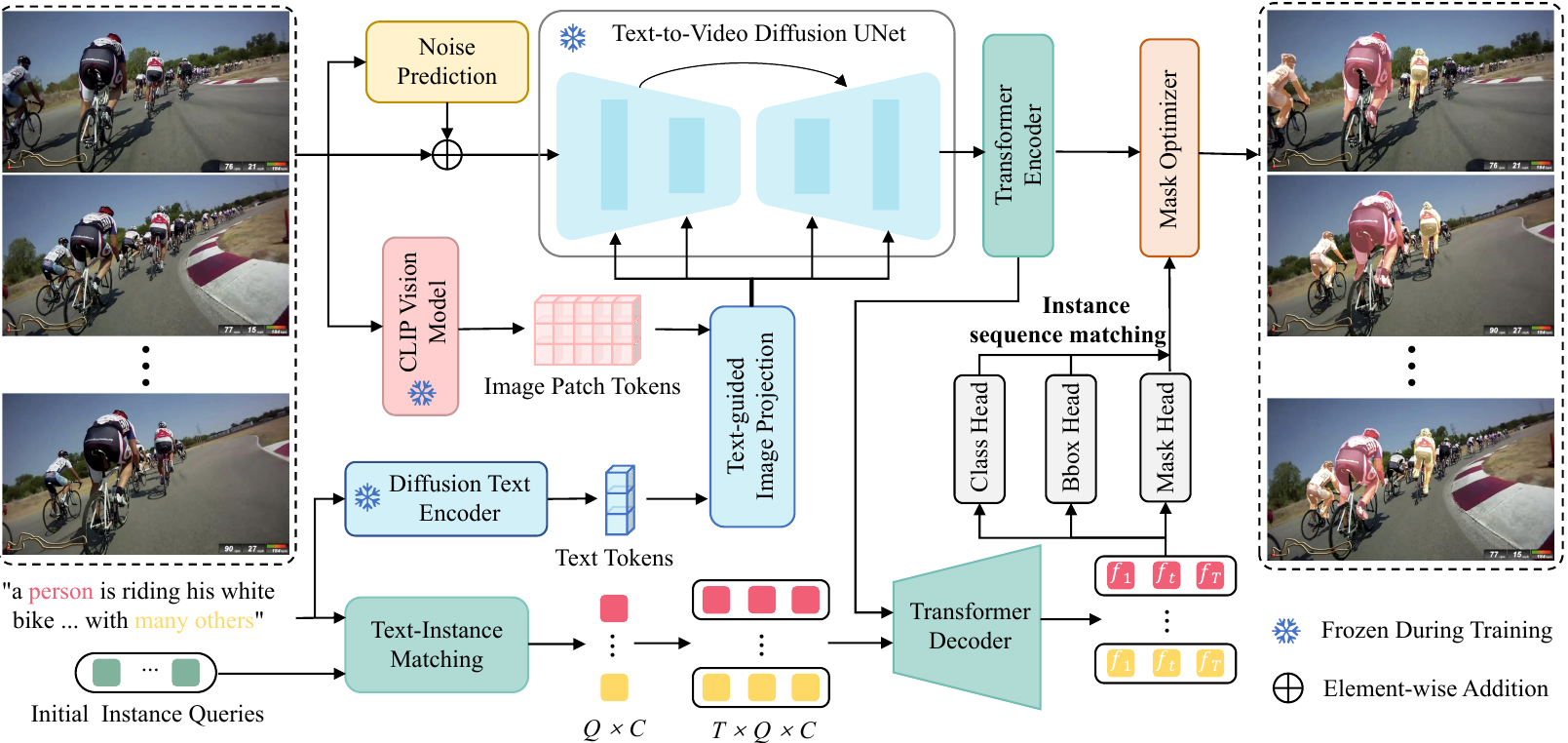}
  \caption{This framework comprises two core components: visual feature extraction and the mask segmentation head. The feature extraction progresses through three phases: (1) prompt generation via text-guided image projection conditions the text-to-video diffusion model; (2), predicted noise is applied to the video; (3) this noisy video and prompt are processed by a diffusion U-Net for visual feature extraction. The segmentation head generates instance queries from the text and merges them with the U-Net's features to create the final masks.
  }
  \label{fig:pp}
   \vspace{-0.5cm}
\end{figure}

By analyzing the task characteristics of R-VOS, we propose a new framework named VD-IT, constructed upon ModelScopeT2V. Illustrated in Figure~\ref{fig:pp}, VD-IT initially generates visual features utilizing the fixed pretrained text-to-video diffusion model, which are subsequently fed into the segmentation head for generating final object masks. As detailed earlier, VD-IT operates as a T2V diffusion model and requires two inputs for extracting the visual representation: a prompt as condition embedding and a noise-added video clip. In the below section, we will describe the corresponding designs in details.

\vspace{-0.5em}
\subsection{Video Feature Extraction}
\vspace{-0.5em}
As introduced in Section~\ref{pre-vre}, extracting diffusion features requires two inputs: the text prompt embedding as condition and the noise-added video. We use Text-Guided Image Projection to generate condition prompt embedding and Noise Prediction to predict the added noise. The methodologies for these two components are elaborated below.

\begin{figure}[tb]
  \centering
  \includegraphics[width=\textwidth]{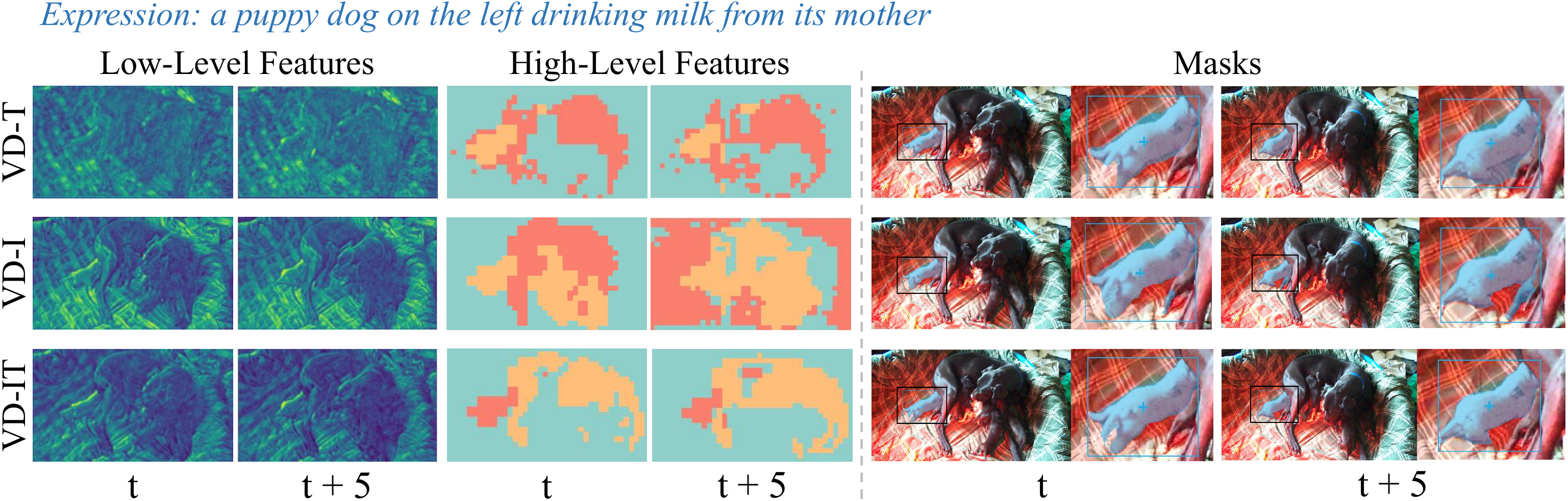}
  \caption{``VD-I'' denotes Image-conditioned Video Diffusion based on the video clip. ``VD-T'' denotes Text-conditioned Video Diffusion based on the referring text, \textit{i.e.}, expression. ``VD-IT'' denotes Image-Text-conditioned Video Diffusion based on both video clip and referring text.
  }
  \vspace{-0.5cm}
  \label{fig:mvit}
\end{figure}

\noindent\textbf{Text-Guided Image Projection.}\label{method:IT}
We investigate three distinct approaches for condition prompt embedding. The first straightforward way is directly using the referring text as the conditional prompt, denoted as ``VD-T''. However, we observed that the resulting segmentation masks often lacked fine-grained details. Illustrated in Figure~\ref{fig:mvit}, despite identifying the target object, the mask quality around the object boundary is not precise enough. We hypothesize that the text prompt serving as a generalized condition and fails to furnish detailed instance-specific information of the U-Net feature for each frame. To validate our guess, we visualize the low-level features of VD-T in Figure~\ref{fig:mvit} and show that these features indeed lack detail.
Inspired by \cite{xu2023open,zhang2024tale}, we further try using the CLIP vision model to extract the visual tokens of each frame followed by a learnable projection layer as an implicit text embedding, termed ``VD-I''. Through further visualization, we find it can preserve the fine-grained details in low-level visual features. However, it will cause some confusion in differentiating object instances in high-level features. As demonstrated in Figure~\ref{fig:mvit}, the final mask erroneously incorporates elements of the background larger dog. We guess this confusion stems from the ignorance of referring text in feature extraction, thus introducing semantic noise into the features and increasing the risk of inaccurately including multiple instances during segmentation. 

Based on the above analysis, we propose to use both referring text tokens and visual tokens as prompts (our ``VD-IT''). As shown in Figure~\ref{fig:mvit}, VD-IT exhibits richer image details in its low-level features compared to VD-T, which leads to more precise mask boundaries. For high-level features, VD-IT demonstrates better temporally aligned semantic visual features with the query text compared to VD-I, enabling accurate temporal matching of subjects, such as a puppy.

More specifically, to extract visual tokens from video frames, we extract vectors using the CLIP~\cite{radford2021learning} vision model to get embedding vectors $\{p_{v,t}\}_{t=1}^{T}, p_{v,t}\in\mathbb{R}^{N_p \times C}$, where $N_p$ denotes the number of tokens and $C$ denotes the channel dimension. For referring text tokens, we utilize the text encoder provided by the text-to-video diffusion model to get the vectors $ p_{e}\in\mathbb{R}^{L \times C}$. We use text token vectors guided image token projection as final prompt, which is formulated by
\begin{equation}
    p_{ve,t} = \text{MLP}(p_{e} + \text{SoftMax}\left(\frac{{p_{e}W^Q}({p_{v,t}W^K})^T}{\sqrt{d_k}}\right){p_{v,t}W^V}),
\end{equation}
where $W^Q, W^K, W^V \in \mathbb{R}^{C \times d_k}$ are learnable parameters, $d_k$ is the number of channels and MLP denotes Multi-Layer Perception. The text tokens play the role of \textit{query} in attention mechanism, and image tokens play the role of \textit{key} and \textit{value}. Therefore, the final prompt combining image and text tokens ensures that the text-to-video diffusion model generates features with both temporal semantic consistency and detailed accuracy.

\vspace{0.5em}
\noindent\textbf{{Noise Prediction.}}
As described in Section~\ref{preliminary}, for a video clip $\mathcal{I}=\left\{I_t\right\}_{t=1}^T$, a text-to-video diffusion model first projects the clip into a latent space using an autoencoder. Then, it proceeds with the diffusion forward pass based on the latent vectors $\mathcal{F}_o = \{f_{o,t}\}_{t=1}^T, f_o\in\mathbb{R}^{\frac{H}{8} \times \frac{W}{8} \times 4}$. During the forward process, the diffusion model will gradually introduce Gaussian noise to the original video signals $\mathcal{F}_o$ based on the $step$. The higher the $step$, the more intense the noise. To maximize the preservation of the original video's visual signals in VD-IT, we set $step = 0$, thereby adding the least amount of noise. Moreover, to avoid blurring key details by using Gaussian noise like \cite{xu2023open,zhang2024tale}, we propose to predict video-correlated noise in place of independent Gaussian noise to retain more details in the features. Specifically, we first send latent vectors  $\mathcal{F}_o$ into a convolution layer and get $\mathcal{F}_n = \{f_{n,t}\}_{t=1}^T, f_n\in\mathbb{R}^{\frac{H}{8} \times \frac{W}{8} \times 4}$.  The prediction of normalized noise $\mathcal{N} = \{n_t\}_{t=1}^{T}, n_t\in\mathbb{R}^{\frac{H}{8} \times \frac{W}{8} \times 4}$ is formulated by
\begin{equation}
    n_t = \left(f_{n,t}W^N - \mu({f_{n,t}W^N}) \right) / \left( \sigma({f_{n,t}W^N}) + \epsilon \right)
\end{equation}
where \(\mu\) and \(\sigma\) represent the mean and standard deviation, respectively, and $W^N\in\mathbb{R}^{4\times4}$ is the trainable weight. \(\epsilon\) is a small constant to prevent division by zero and set as 1e-5 in our experiments by default.

With the prompts $p_{ve,t}$ and predicted noise $n_t$ fed into the pretrained text-to-video model, we extract side outputs from each intermediate layer of the diffusion U-Net as visual features $\mathcal{F}_v^D$. The formulation is 
\begin{equation}
    \mathcal{F}_v^D = \text{U-Net}(\alpha_0\mathcal{F}_o + (1-\alpha_0)\mathcal{N}, \{p_{ve,t}\}_{t=1}^T),
\end{equation}
where $1 - \alpha_0$ denotes the noise strength of the $0$-th step, determined by the noise schedule inherent to the diffusion model. $\mathcal{F}_v^D$ includes 3-level visual features downsampled by factors of $8\times$ to $32\times$.  By incorporating features downsampled by $4\times$ from the intermediate layer of the autoencoder, we construct a 4-level hierarchy of visual features $\mathcal{F}_v$, with downsampling rates from $4\times$ to $32\times$.
\vspace{-0.5em}
\subsection{Mask Prediction Head}
\vspace{-0.5em}
After obtaining the video feature representations, we follow the widely used query-based segmentation model design and propose a simple mask prediction head that first extracts instance queries from referring text tokens and then fuse there queries with visual features to produce the final segmentation masks.

\noindent\textbf{Text-Instance Matching.} The mask prediction head first extracts instance queries from referring text tokens. Here, we first have $Q$ learnable query instance vectors $q_o\in\mathbb{R}^{Q \times C}$ which are  randomly initialized, where Q represents the number of instance queries. These instance vectors undergo cross-attention with word vectors $\mathcal{F}_e = \{f_{e,l}\}_{l=1}^L, f_{e,l}\in\mathbb{R}^{C}$ extracted from the referring text  by RoBERTa~\cite{liu2019roberta}. In detail, $q_o$ will serve as the \textit{query}, while $\mathcal{F}_e$ serves as both the \textit{key} and \textit{value}. The updated query instance vectors $q_e\in\mathbb{R}^{Q \times C}$ then represent the different objects described within the referring expression. 

\noindent\textbf{Vision-Language Fusion.} Then the segmentation head fuses instance queries and visual features to output final masks.  Following the common practice~\cite{miao2023spectrum,wu2022language,botach2022end}, we feed the visual features $\mathcal{F}_v$ into a Deformable Transformer\cite{zhu2020deformable} encoder to deal with multi-scale features. Next, we use them as the \textit{key} and \textit{value} in a Deformable Transformer decoder. Concurrently, instance query vectors $q_e\in\mathbb{R}^{Q \times C}$ are replicated $T$ times along the temporal dimension and introduced into the transformer decoder as \textit{query}, resulting in the extraction of cross-modality features $\mathcal{F}_{ve}\in\mathbb{R}^{T \times \frac{H}{8} \times \frac{W}{8} \times C_d}$.

\noindent\textbf{Segmentation Head.} With the cross-modality features $\mathcal{F}_{ve}$, the segmentation head sends it into three sub-heads: bounding box head, classification head and mask head. We configure the bounding box head as a 3-layer feed-forward network (FFN) with ReLU activation for all except the final layer, while the classification head consists of a linear layer. These two heads process $\mathcal{F}_{ve}$, to produce final bounding box predictions, $\mathcal{B}\in\mathbb{R}^{T \times Q \times 4}$ and classification predictions, $\mathcal{C}\in\mathbb{R}^{T \times Q}$. The classification head functions as a binary classifier, determining whether each frame contains the object of interest. Concurrently, dynamic convolution is performed between $\mathcal{F}_{ve}$ and $\mathcal{F}_{v}$ to produce the segmented masks $\mathcal{M}_o\in\mathbb{R}^{T \times Q \times \frac{H}{8} \times \frac{W}{8}}$. Finally, the predicted masks are enlarged $8\times$ through four times of upsampling to reach final masks $\mathcal{M}\in\mathbb{R}^{T \times Q \times H \times W}$. Before the first and third rounds of upsampling, they are combined with visual features $\mathcal{F}_{v}$ to achieve high-quality masks.

To accurately identify each referred object in videos, we use Hungarian algorithm~\cite{Kuhn1955TheHM} to select the most suitable match from $Q$ instance queries. For the matched instance, We adopt the same training losses and weights as used in~\cite{miao2023spectrum,wu2022language,botach2022end} for a fair comparison. Specifically, we use Dice loss~\cite{li2019dice} and Focal loss~\cite{lin2017focal} for mask $\mathcal{M}$, Focal loss~\cite{lin2017focal} for confidence scores $\mathcal{S}$, and L1 and GIoU~\cite{rezatofighi2019generalized} loss for bounding boxes $\mathcal{B}$. In inference, we choose the instance query which has the high confidence scores $\mathcal{S}$ as our final results.

\vspace{-0.5em}
\section{Experiments}
\vspace{-0.5em}

\noindent\textbf{Datasets and Evaluation Metrics.} We conduct evaluations on four popular R-VOS benchmarks. Ref-Youtube-VOS~\cite{seo2020urvos} is the largest, with 3,978 videos and 15,000 language descriptions. Ref-DAVIS17~\cite{khoreva2019video} includes descriptions for specific objects in 90 videos. A2D-Sentences~\cite{gavrilyuk2018actor} has 3,782 videos and each video is annotated with 3-5 frames, totaling over 6,600 annotations. JHMDB-Sentences~\cite{gavrilyuk2018actor} has 928 videos and corresponding descriptions, covering 21 action classes. On Ref-Youtube-VOS and Ref-DAVIS17, the performance is measured using standard metrics: region similarity ($\mathcal{J}$), contour accuracy ($\mathcal{F}$), and their mean ($\mathcal{J} \& \mathcal{F}$). We also report the frames per second (FPS) achieved on Ref-Youtube-VOS. For A2D-Sentences and JHMDB-Sentences, we employ Overall IoU, Mean IoU, and mAP within interval of 0.50 to 0.95, incremented by 0.05.

\noindent\textbf{Training Details.} Following common practices, we train our models on the Ref-YouTube-VOS training set and test them on the validation splits of Ref-YouTube-VOS and Ref-DAVIS17. Since some methods~\cite{miao2023spectrum, wu2022language} report their results after pre-training on Ref-COCO~\cite{yu2016modeling}, we also report our results on this setting followed by fine-tuning on Ref-YouTube-VOS. The pre-trained model is also fine-tuned on the A2D-Sentences training set, and we then evaluate its performance on the validation splits of A2D-Sentences and JHMDB-Sentences. The number $Q$ of queries across all datasets is set to 5. The training of the models utilizes two NVIDIA A100 GPUs, processing 5 frames per clip over the course of 9 epochs. To accommodate model requirements, all frames are adjusted to ensure the longest side to be 640 pixels. Additional training details can be found in the supplementary materials.

\vspace{-0.5em}
\subsection{Comparison with State-of-the-Art Methods}
\vspace{-0.5em}
\noindent\textbf{Ref-YouTube-VOS and Ref-DAVIS17.} We compare our method with other state-of-the-art methods in Table~\ref{tab:sotatv}. It can be seen that our approach outperforms current methods on both datasets for all measured metrics. Specifically, on Ref-YouTube-VOS, VD-IT achieves 64.8 $\mathcal{J} \& \mathcal{F}$ which is 3.2 point higher than the previous state-of-the-art SgMg~\cite{miao2023spectrum}. This indicates the effectiveness of our model in maintaining consistent performance over time and producing high-quality masks, leading to an overall improvement in segmentation results. When pre-training with RefCOCO/+/g, on the Ref-DAVIS17 dataset, VD-IT records a $\mathcal{J} \& \mathcal{F}$ score of 69.4, surpassing the previous state-of-the-art by 6.1. This significant improvement underscores the versatility and effectiveness of our approach. 

\begin{table}[t]
\centering
\caption{Comparison with the state-of-the-art methods on Ref-Youtube-VOS and Ref-DAVIS17. * denotes that we run the official codes to get the results.}
\label{tab:sotatv}
\vspace{-0.3cm}
\footnotesize
\setlength{\tabcolsep}{1.0mm}
\begin{tabularx}{\textwidth}{l|c|*{4}{C}|*{3}{C}}
\hline \multirow{2}{*}{Method} & \multirow{2}{*}{Backbone} & \multicolumn{4}{c|}{Ref-YouTube-VOS} & \multicolumn{3}{c}{Ref-DAVIS17} \\
\cline{3-9} & & $\mathcal{J} \& \mathcal{F}$ & $\mathcal{J}$ & $\mathcal{F}$ & FPS & $\mathcal{J} \& \mathcal{F}$ & $\mathcal{J}$ & $\mathcal{F}$ \\
\hline
 CMSA~\cite{ye2019cross} & ResNet-50 & 36.4 & 34.8 & 38.1 & - & 40.2 & 36.9 & 43.5 \\
 URVOS~\cite{seo2020urvos} & ResNet-50 & 47.2 & 45.3 & 49.2 & - & 51.5 & 47.3 & 56.0 \\
 CMPC-V~\cite{liu2021cross} & I3D & 47.5 & 45.6 & 49.3 & - & - & - & - \\
 PMINet~\cite{ding2021progressive} & ResNeSt-101 & 53.0 & 51.5 & 54.5 & - & - & - & - \\
 YOFO~\cite{li2022you} & ResNet-50 & 48.6 & 47.5 & 49.7 & 10 & 53.3 & 48.8 & 57.8 \\
 LBDT~\cite{ding2022language} & ResNet-50 & 49.4 & 48.2 & 50.6 & - & 54.3 & - & - \\
 MLRL~\cite{wu2022multi} & ResNet-50 & 49.7 & 48.4 & 51.0 & - & 52.8 & 50.0 & 55.4 \\
 MTTR~\cite{botach2022end} & Video-Swin-T & 55.3 & 54.0 & 56.6 & - & - & - & - \\
 MANet~\cite{chen2022multi} & Video-Swin-T & 55.6 & 54.8 & 56.5 & - & - & - & - \\
ReferFormer~\cite{wu2022language}& Video-Swin-T & 56.0 & 54.8 & 57.3 & 50 & - & - & - \\
SgMg~\cite{miao2023spectrum}  & Video-Swin-T & 58.9 & 57.7 & 60.0 & 65 & 56.7 & 53.3 & 60.0 \\
SgMg*~\cite{miao2023spectrum}  & Video-Swin-B & 61.6 & 59.7 & 63.5 & 40 & - & - & - \\
VD-IT (Ours) & Video Diffusion & $\mathbf{64.8}$ & $\mathbf{63.1}$ & $\mathbf{66.6}$ & 21 & $\mathbf{63.0}$ & $\mathbf{59.9}$ & $\mathbf{66.1}$\\
 \hline
\multicolumn{9}{c}{Pre-training with RefCOCO/+/g} \\
\hline
 ReferFormer~\cite{wu2022language} & Video-Swin-T & 59.4 & 58.0 & 60.9 & 50 & 59.6 & 56.5 & 62.7 \\
 SgMg~\cite{miao2023spectrum}  & Video-Swin-T & 62.0 & 60.4 & 63.5 & 65 & 61.9 & 59.0 & 64.8 \\
 ReferFormer~\cite{wu2022language} & Video-Swin-B & 62.9 & 61.3 & 64.6 & 33 & 61.1 & 58.1 & 64.1 \\
 OnlineRefer~\cite{wu2023onlinerefer} & Video-Swin-B & 62.9 & 61.0 & 64.7 & - & 62.4 & 59.1 & 65.6 \\
 SgMg~\cite{miao2023spectrum}  & Video-Swin-B & 65.7 & 63.9 & 67.4 & 40 & 63.3 & 60.6 & 66.0 \\
 VD-IT (Ours)  & Video Diffusion & $\mathbf{66.5}$ & $\mathbf{64.4}$ & $\mathbf{68.5}$ & 21 & $\mathbf{69.4}$ & $\mathbf{66.2}$ & $\mathbf{72.6}$\\
\hline
\end{tabularx}
\vspace{-0.3cm}
\end{table}

\begin{table}
\centering
\caption{Quantitative comparison to state-of-the-art R-VOS methods on A2D-Sentences and JHMDB-Sentences.}
\vspace{-0.3cm}
\label{tab:sotaah}
\footnotesize
\begin{tabularx}{\textwidth}{l|c|*{3}{C}|*{3}{C}}
\hline \multirow{2}{*}{ Method } & \multirow{2}{*}{ Backbone } & \multicolumn{3}{|c|}{ A2D-Sentences } & \multicolumn{3}{|c}{ JHMDB-Sentences } \\
& & mAP & Overall & Mean & mAP & Overall  & Mean \\
\hline Hu et al.~\cite{hu2016segmentation} & VGG-16 & 13.2 & 47.4 & 35.0 & 17.8 & 54.6 & 52.8 \\
Gavrilyuk et al.~\cite{gavrilyuk2018actor} & I3D & 19.8 & 53.6 & 42.1 & 23.3 & 54.1 & 54.2 \\
ACAN~\cite{wang2019asymmetric} & I3D & 27.4 & 60.1 & 49.0 & 28.9 & 57.6 & 58.4 \\
CMPC-V~\cite{liu2021cross} & I3D & 40.4 & 65.3 & 57.3 & 34.2 & 61.6 & 61.7 \\
ClawCraneNet~\cite{liang2021clawcranenet} & ResNet-50/101 & - & 63.1 & 59.9 & - & 64.4 & 65.6 \\
MTTR~\cite{botach2022end} & Video-Swin-T & 46.1 & 72.0 & 64.0 & 39.2 & 70.1 & 69.8 \\
ReferFormer~\cite{wu2022language} & Video-Swin-T & 52.8 & 77.6 & 69.6 & 42.2 & 71.9 & 71.0 \\
SgMg~\cite{miao2023spectrum}  & Video-Swin-T & 56.1 & 78.0 & 70.4 & 44.4 & 72.8 & 71.7 \\
ReferFormer~\cite{wu2022language} & Video-Swin-B & 55.0 & 78.6 & 70.3 & 43.7 & 73.0 & 71.8 \\
SgMg~\cite{miao2023spectrum} & Video-Swin-B & 58.5 & 79.9 & 72.0 & 45.0 & 73.7 & 72.5 \\
VD-IT (Ours) & Video Diffusion & $\mathbf{61.4}$ & $\mathbf{81.5}$ & $\mathbf{73.2}$ & $\mathbf{46.5}$ & $\mathbf{74. 4}$ & $\mathbf{73. 4}$ \\
\hline
\end{tabularx}
\vspace{-0.3cm}
\end{table}

\noindent\textbf{A2D-Sentences and JHMDB-Sentences.} We further evaluate our method on the A2D-Sentences and JHMDB-Sentences datasets and compare its performance with other state-of-the-art methods in Table~\ref{tab:sotaah}. In alignment with methodologies from recent studies~\cite{miao2023spectrum,wu2022language}, the approach involves pre-training on the RefCOCO/+/g datasets before fine-tuning on A2D-Sentences, while JHMDB-Sentences is exclusively used for evaluation purposes. It shows that VD-IT outperforms state-of-the-art methods by 2.9 and 1.5 in mAP for A2D-Sentences and JHMDB-Sentences, respectively. This performance advantages highlights the model's enhanced performance on datasets characterized by extensive action variations, showcasing a greater robustness to camera and object movements compared to conventional methods. 

\subsection{Ablation Study}
In this section, we perform extensive ablation studies on
Ref-Youtube-VOS to study the effect of different components in
our model. ``Image-Cond'' and ``Text-Cond'' shown in Table~\ref{tab:abdesign} denotes only using visual token or text token in VD-I and VD-T, while ``NP'' denotes Noise Prediction.

\begin{table}[t]
\centering
\begin{minipage}{0.49\textwidth}
\centering
\caption{Ablation study on the components of VD-IT on Ref-YouTube-VOS. ``NP'' denotes Noise Prediction.}
\vspace{-0.3cm}
\label{tab:abdesign}
\footnotesize
\begin{tabularx}{\textwidth}{*{3}{C}  *{3}{C}}
\hline \multicolumn{3}{c}{ Components } & \multicolumn{3}{c}{ Performance } \\
\hline
Image-Cond & Text-Cond & \multirow{2}{*}{NP} & \multirow{2}{*}{$\mathcal{J} \& \mathcal{F}$} & \multirow{2}{*}{$\mathcal{J}$} & \multirow{2}{*}{$\mathcal{F}$} \\
\hline
$\checkmark$ & & & 59.7 & 57.9 & 61.6  \\
& $\checkmark$ & & 61.9 & 60.1 & 63.7  \\
$\checkmark$ & $\checkmark$ & & 63.8 & 62.0 & 65.5\\
$\checkmark$ & $\checkmark$ & $\checkmark$ & $\mathbf{64.8}$ & $\mathbf{63.1}$ & $\mathbf{66.6}$ \\
\hline
\end{tabularx}
\end{minipage}%
\hfill
\begin{minipage}{0.49\textwidth}
\centering
\caption{Quantitative evaluation on the validation split of RefCOCO/+/g. Overall IoU is adopted as the evaluation metric.}
\vspace{-0.3cm}
\label{tab:sotaimg}
\footnotesize
\begin{tabularx}{\textwidth}{l|C|C|C}
\hline Method & Ref. & Ref.+ & Ref.g \\
\hline MaIL~\cite{li2021mail} & 70.1 & 62.2 & 62.5 \\
CRIS~\cite{wang2022cris} & 70.5 & 62.3 & 59.9 \\
RefTR~\cite{li2021referring} & 70.6 & - & - \\
LAVT~\cite{yang2022lavt} & 72.7 & 62.1 & 61.2 \\
VLT~\cite{ding2022vlt} & 73.0 & 63.5 & 63.5 \\
SgMg~\cite{miao2023spectrum} & 76.3 & 66.4 & 70.0\\
VD-IT (Ours) &  $\mathbf{76.7}$ &  $\mathbf{66.5}$ &  $\mathbf{70.3}$\\
\hline
\end{tabularx}
\end{minipage}
\vspace{-0.5cm}
\end{table}

\vspace{0.5em}
\noindent\textbf{Component Analysis.} As described in Section~\ref{method:IT}, we have tried three different settings for the prompt embedding to extract the feature representation and the corresponding ablation results are shown in Table~\ref{tab:abdesign}. By comparing the results with the state-of-the-art results shown in Table~\ref{tab:sotatv}, we can observe that just using visual tokens as the condition embedding like \cite{xu2023open,zhang2024tale} cannot perform as well as SgMg~\cite{miao2023spectrum} that uses the discriminative Video Swin model as the backbone. This also indicates that our state-of-the-art (SOTA) performance stems neither from increased capacity nor from more extensive data exposure in the text-to-video diffusion model. 
By considering temporal semantics provided by the referring text, VD-T achieves a stronger performance with 2.2 point improvement. This supports our initial hypothesis that referring text plays a crucial role in enhancing visual features. The oversight of referring text in the visual encoders of previous methods is not optimal. However, relying solely on referring text for conditioning captures only temporal semantic consistency without extracting detailed visual features. By leveraging image tokens for detail alongside text tokens for semantic guidance, our final model sees a large performance boost. Besides the prompt embedding choice, in the forward process, we choose to predict video-specific noise instead of using normal Gaussian noise, further boosting the performance by 1.0 point.

\begin{table}[t]
\centering
\begin{minipage}{0.45\textwidth}
\centering
\caption{Temporal consistency ablation. 
"IoU diff." denotes the IoU variation between frames and "Inter." denotes the frame interval. Averages are calculated on Ref-YouTube-VOS.}
\vspace{-0.3cm}
\label{tab:abtc}
\footnotesize
\begin{tabularx}{\textwidth}{l|C|C}
\hline\multirow{2}{*}{ Method}  & \multicolumn{2}{c}{ IoU Diff.} \\
\cline{2-3} & Inter. 1  & Inter. 5 \\
\hline
SgMg~\cite{miao2023spectrum} & 7.24 & 11.15 \\
VD-I & 6.52 & 9.43 \\
VD-T  & 5.41 & 8.25\\
VD-IT & $\mathbf{5.19}$ & $\mathbf{7.89}$\\
\hline
\end{tabularx}
\end{minipage}%
\hfill
\begin{minipage}{0.5\textwidth}
\centering
\caption{
Ablation study shows the impact of fine-grained features on mask quality in Ref-YouTube-VOS, reporting the ratio of high-quality (IoU > 0.9) to moderate-quality (IoU > 0.5) masks.}
\vspace{-0.3cm}
\label{tab:tahq}
\footnotesize
\begin{tabularx}{\textwidth}{*{3}{C} | c}
\hline \multicolumn{3}{c|}{ Components } & \multirow{3}{*}{$\text{Ratio} = \frac{\text{IoU} > 0.9}{\text{IoU} > 0.5}$} \\
\cline{1-3}
Image-Cond & Text-Cond  & \multirow{2}{*}{NP} & \\
\hline
$\checkmark$ & & & 47.26\%  \\
& $\checkmark$ & & 44.54\%  \\
$\checkmark$ & $\checkmark$ & & 46.71\%\\
$\checkmark$ & $\checkmark$ & $\checkmark$ & $\mathbf{47.36\%}$ \\
\hline
\end{tabularx}
\end{minipage}
\vspace{-0.4cm}
\end{table}

\vspace{0.5em}
\noindent\textbf{Temporal Consistency Analysis.}
To demonstrate that our VD-IT offer better temporal consistency than V-Swin as backbones, Table~\ref{tab:abtc} presents the difference in IoU between two frames, \textit{i.e.}, frame interval is 1 and 5. A smaller difference in IoU indicates improved temporal coherence. When the visual encoder does not incorporate referring text, the distinction between VD-I and SgMg becomes less significant. The minor advantage observed with VD-I is primarily due to its enhanced denoising capabilities, which improve resilience to various types of noise such as changes in lighting and camera movements. Consequently, the fluctuation in IoU scores for VD-I shows a reduction of about 0.7 compared to SgMg. reflecting a steadier performance across diverse scenarios. 
By incorporating referring text within the visual encoder to guide feature extraction,  VD-T and VD-IT exhibit much improved temporal consistency. Their fluctuation in performance measures is around 2 point less than that observed with SgMg. This reduction in variability demonstrates that VD-IT significantly enhances performance through better maintenance of temporal consistency.

\vspace{0.5em}
\noindent\textbf{Enhancing Mask Quality.}
To further study our design in enhancing the model to generate high-quality masks, we use the ratio of masks with IoU greater than 0.9 to those with IoU above 0.5 as the metric. A higher ratio reflects a better capacity for generating more high-quality masks. As shown in Table~\ref{tab:tahq}, VD-I, when utilizing visual tokens as prompts, provides additional details to visual features compared to VD-T, which relies solely on referring text for prompts. Compared to only using visual tokens, combining referring text tokens with visual tokens will slightly worsen the high-quality mask generation. But by using our predicting noise module, our VD-IT can extract finer features and produce slightly higher-quality masks. This ablation analysis underscores the significance of integrating detailed visual information through image tokens and the nuanced improvements afforded by noise prediction in refining the feature extraction process for superior mask generation.

\begin{figure}[t]
    \centering
\begin{minipage}[t]{0.48\textwidth}
\includegraphics[width=\linewidth]{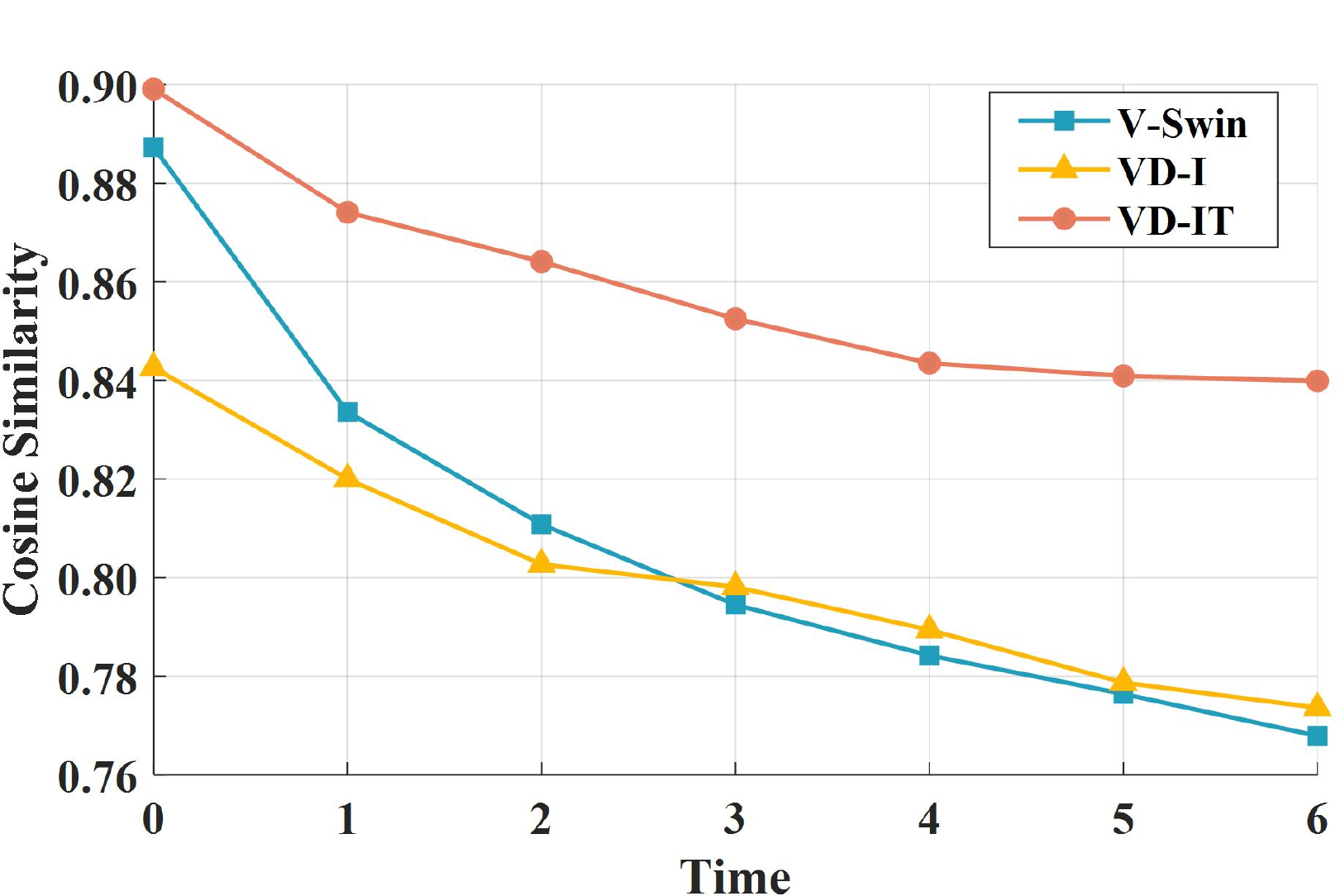}
\captionof{figure}{Temporal Semantic Consistency. Averaged over 1,000 samples from Ref-Youtube-VOS, the cosine similarity between the Region of Interest (RoI) features of the initial frame and the following seven frames is reported.}
\label{fig:mv2}
\end{minipage}%
\hfill 
\begin{minipage}[t]{0.48\textwidth}
  \centering
\includegraphics[width=0.8\linewidth]{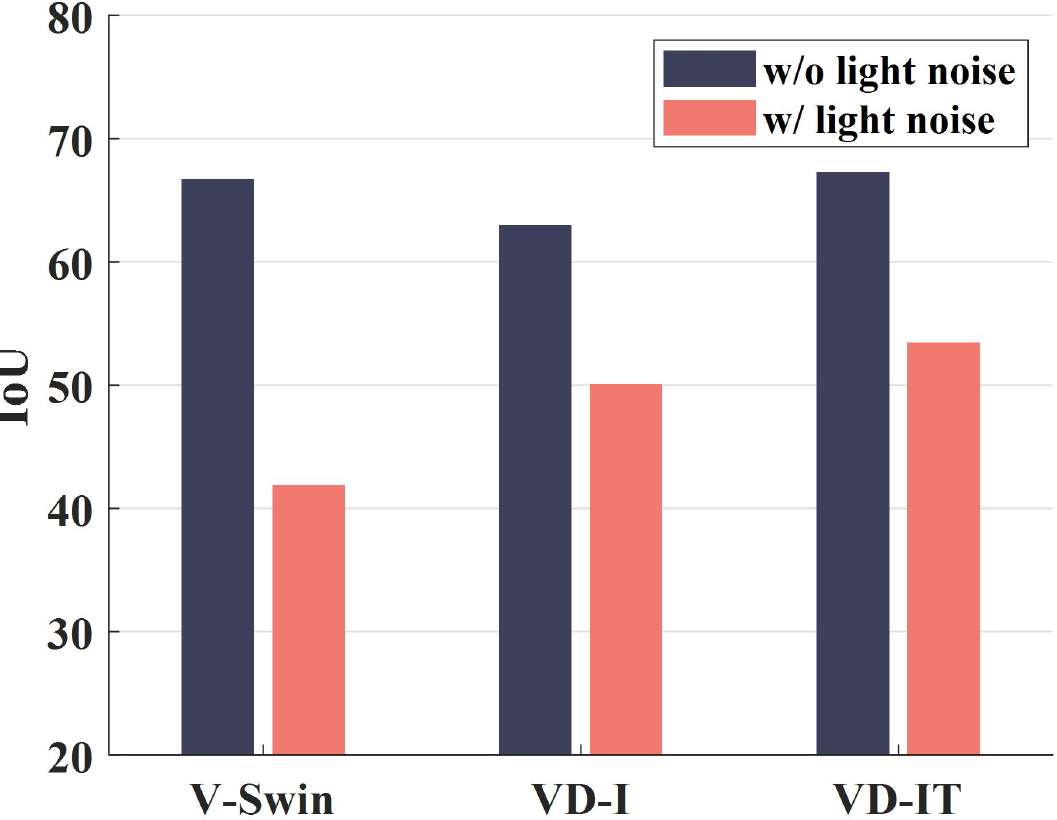}
\captionof{figure}{Robustness against light noise. We modify the brightness of various frames randomly and compare the IoU of segmentation results under changing lighting conditions. The results are reported on Ref-Youtube-VOS.}
\label{fig:lnoise}
\end{minipage}
\vspace{-0.5cm}
\end{figure}

\vspace{0.5em}
\noindent\textbf{Discussion on Feature Behavior.}\label{lightex}
In our previous experiments, we validated the good temporal consistency of the T2V model through an analysis of the segmentation results. However, we did not confirm whether it was indeed the prompt tuning that enabled the model to learn features with strong temporal semantic consistency, leading to temporally coherent outcomes. Therefore, we further test the temporal semantic similarity of the features. Figure~\ref{fig:mv2} shows that both VD-I and VD-IT have a slower semantic decline, a phenomenon that visual features would gradually drift over consecutive video frames, compared to the V-Swin, indicating better temporal stability. Importantly, VD-IT benefits from using referring text for semantic guidance, which results in higher semantic similarity than VD-I. We hypothesize that the slower decline of VD-I and VD-It compared to V-Swin is due to their stronger robustness to environmental variations in videos. Therefore, we evaluated the model performance under varying lighting conditions, as depicted in Figure~\ref{fig:lnoise}. To make the results more convincing, we select SgMg~\cite{miao2023spectrum}, a state-of-the-art method that previously utilized V-Swin as the visual encoder, for comparison with VD-I and VD-IT. Figure~\ref{fig:lnoise} shows that VD-I and VD-IT are more resistant to lighting noise, explaining their slower rate of semantic decline.

\begin{figure}[tb]
  \centering
  \includegraphics[width=\textwidth]{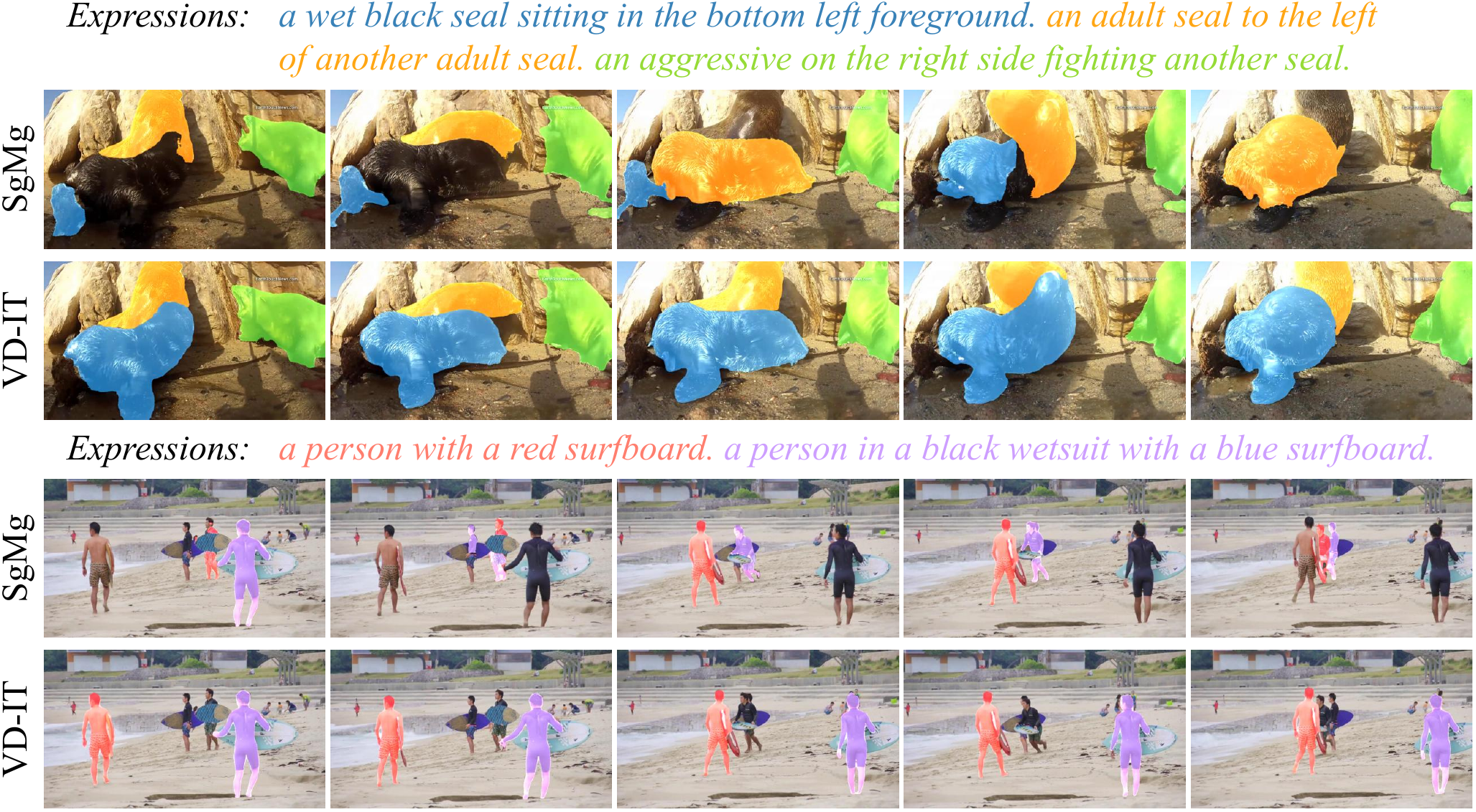}
  \caption{Qualitative comparison of our method with others. VD-IT has better temporal consistency.
  }
  \label{fig:visualresults}
  \vspace{-0.5cm}
\end{figure}

\vspace{0.5em}
\noindent\textbf{Referring Image Segmentation Results.} We implemented VD-IT for referring image segmentation directly, comparing it to the current state-of-the-art methods on the RefCOCO/+/g datasets, without making any changes to the architecture. According to the results shown in Table~\ref{tab:sotaimg}, VD-IT achieves strong performance on all three datasets. Nonetheless, its performance is quite similar to that of SgMg, which highlights our progress on video R-VOS datasets due to our focus on maintaining temporal consistency, not just on improving single-frame image mask quality. If improving single-frame mask quality were our main achievement, we would expect much higher IoU scores on RefCOCO datasets compared to SgMg. This suggests our method's strength lies in enhancing performance through attention to time-based continuity rather than simply boosting the quality of individual image masks. 

\vspace{0.5em}
\noindent\textbf{Visualization Results.} In Figure~\ref{fig:visualresults}, we further provide some visual comparison results with the state-of-the-art method SgMg \cite{miao2023spectrum}. 
From the visual results, we can clearly see that our proposed VD-IT can segment and track the referred object in a more temporally consistent way than the baseline method.

\section{Conclusion}
This paper presents a pioneering exploration into leveraging feature representations learned in pretrained generative text-to-video diffusion models for video understanding tasks. We take the classical referring video object segmentation task as the testbed, and design a new R-VOS framework with some dedicated designs. Through comprehensive experimentation, we validate the effectiveness of our proposed R-VOS framework, showcasing superior performance over existing state-of-the-art methods. Our findings underscore the capacity of latent representations learned in generative text-to-video models to encapsulate rich semantics and coherent temporal correspondence. Consequently, the segmentation results based on these latent features exhibit markedly improved temporal consistency compared to methods utilizing feature representations of video backbones fine-tuned in a discriminative manner. We envision that our pioneering exploration will inspire further insightful research in this direction, propelling the unification of generative and discriminative models toward new horizons.

%
%
\bibliographystyle{splncs04}
\bibliography{main}

\appendix
\newpage

\section{Additional Implementation Details}

\noindent\textbf{Additional Training Details.} In the pre-training phase on Ref-COCO~\cite{yu2016modeling}, we initiate with distinct learning rates for different components: 2.5e-6 for the text encoder, and 2.5e-5 for the remaining parts of the model. This stage spans 12 epochs, focusing on single-frame inputs. The learning rates undergo a reduction by a factor of 10 at the 8-th and 10-th epochs to refine the training process. In our main training phase, we adjust the training length to 6 or 9 epochs, depending on whether we use pre-training or not. The text encoder is kept frozen during training. Initial learning rates are set at 5e-5 for the whole model. For a 9-epoch training setting, learning rates are decreased by a factor of 10 at the $6$-th and $8$-th epochs. In contrast, for a 6-epoch schedule, reductions occur at the $3$-rd and $5$-th epochs. 

\noindent\textbf{Additional Training Loss Details.}
To ensure a fair comparison, we utilize the identical training losses and weights that are mentioned in \cite{miao2023spectrum,wu2022language,botach2022end}. 
Therefore, our segmentation consists of three sub-heads: bounding box head, classification head and mask head. These correspond to three types of outputs:  bounding boxes $\mathcal{B}$, classification scores $\mathcal{S}$, and segmentation masks, which include both a low-resolution ${\mathcal{M}_o}$ and a refined high-resolution ${\mathcal{M}}$. 
The overall training loss equation used to optimize these outputs is 
\begin{equation}
\mathcal{L}_{\text {train }}=\lambda_{\mathcal{M}_o} \mathcal{L}_{\mathcal{M}_o}+\lambda_{\mathcal{M}} \mathcal{L}_{\mathcal{M}}+\lambda_{\mathcal{B}} \mathcal{L}_{\mathcal{B}}+\lambda_{\mathcal{S}} \mathcal{L}_{\mathcal{S}},
\end{equation}
where $\mathcal{L}$ represents the different loss components and $\lambda$ denotes their respective weights. Specifically, $\mathcal{L}_{\mathcal{M}}$ and $\mathcal{L}_{\mathcal{M}_o}$ are mask losses. $\mathcal{L}_{\mathcal{B}}$ is bounding box loss and $\mathcal{L}_{\mathcal{S}}$ is classification loss. Specifically, we use Dice loss~\cite{li2019dice} and Focal loss~\cite{lin2017focal} for masks $\{\mathcal{M}_o, \mathcal{M}\}$, Focal loss~\cite{lin2017focal} for confidence scores $\mathcal{S}$, and L1 and GIoU~\cite{rezatofighi2019generalized} loss for bounding boxes $\mathcal{B}$. 
 
\noindent\textbf{Additional Matching Loss Details.}
Our approach to instance matching is in line with the recent transformer-based methods~\cite{wang2021end,wu2022defense,miao2023spectrum,wu2022language,botach2022end}. In particular, we adopt the Hungarian algorithm~\cite{Kuhn1955TheHM} to select the most suitable match between the $Q=5$ instance queries and the ground truth. For this purpose, final masks ${\mathcal{M}}$, bounding boxes $\mathcal{B}$ and classification scores $\mathcal{S}$ are used to compute the matching loss $\mathcal{L}_{\text {match }}$ for each query and Hungarian algorithm is used to find the best match that has the minimum loss. $\mathcal{L}_{\text {match }}$ lies in three parts, which can be formulated as
\begin{equation}
\mathcal{L}_{\text {match }}=\lambda_{\mathcal{M}} \mathcal{L}_{\mathcal{M}}+\lambda_{\mathcal{B}} \mathcal{L}_{\mathcal{B}}+\lambda_{\mathcal{S}} \mathcal{L}_{\mathcal{S}}.
\end{equation}

\begin{figure}[tb]
  \centering
  \includegraphics[width=\textwidth]{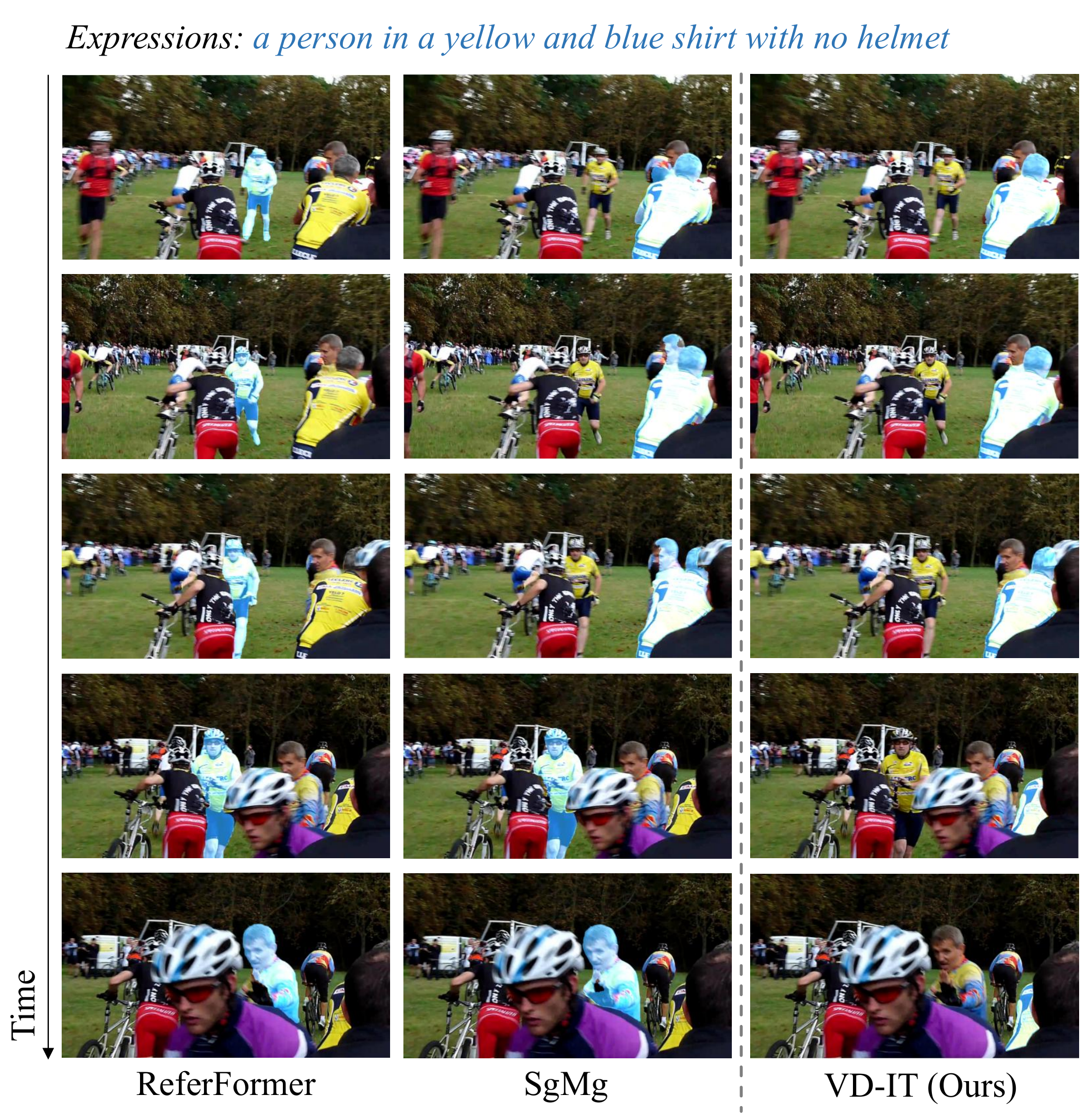}
  \caption{Qualitative comparison of our method with others. Both ReferFormer~\cite{wu2022language} and SgMg~\cite{miao2023spectrum} use V-Swin as the visual encoder. 
  }
  \label{fig:mvr1}
   \vspace{-0.5cm}
\end{figure}

\section{Additional Visualization Results}

To further highlight the benefits of using the T2V diffusion model as the visual encoder over V-Swin, we showcase a comparison between our VD-IT and several V-Swin based methods in Figure~\ref{fig:mvr1}. VD-IT demonstrates superior temporal consistency in tracking individuals, particularly when the person is extensively obscured. This enhancement becomes especially evident in challenging scenarios where maintaining continuity and accuracy in tracking is critical, highlighting VD-IT's robustness in handling complex visual occlusions.

\begin{table}[t]
\centering
\begin{minipage}{0.49\textwidth}
\centering
\caption{Ablation study on the effectiveness of attention mechanism in Text-Guided Image Projection. ``Attn,'' denotes the attention mechanism. ``Concat.'' denotes concatenation. All results are reported on Ref-YouTube-VOS.}
\vspace{-0.3cm}
\label{tab:tim}
\footnotesize
\begin{tabularx}{\textwidth}{l|C|C|C|C}
\hline Method & Fusion & $\mathcal{J} \& \mathcal{F}$ & $\mathcal{J}$ & $\mathcal{F}$ \\
\hline 
VD-I & - & 59.7 & 57.9 & 61.6 \\
VD-T & - & 61.9 & 60.1 & 63.7 \\
VD-IT & Concat. & 62.4 & 60.8 & 64.0 \\
VD-IT & Attn. &  $\mathbf{64.8}$ & $\mathbf{63.1}$ & $\mathbf{66.6}$\\
\hline
\end{tabularx}
\end{minipage}%
\hfill
\begin{minipage}{0.49\textwidth}
\centering
\caption{Quantitative evaluation on the validation split of RefCOCO/+/g. Overall IoU is adopted as the evaluation metric.}
\vspace{-0.3cm}
\label{tab:nstep}
\footnotesize
\begin{tabularx}{\textwidth}{l|C|C|C}
\hline Step & $\mathcal{J} \& \mathcal{F}$ & $\mathcal{J}$ & $\mathcal{F}$ \\
\hline 1 &  $\mathbf{64.8}$ & $\mathbf{63.1}$ & $\mathbf{66.6}$ \\
5 & 63.4 & 61.6 & 65.1 \\
10 & 63.5 & 61.7 & 65.1 \\
50 & 63.1 & 61.3 & 64.9 \\
100 & 62.7 & 60.8 & 64.5\\

\hline
\end{tabularx}
\end{minipage}
\vspace{-0.5cm}
\end{table}

\section{Additional Experiment Results}

\noindent\textbf{Text-Guided Image Projection.} 
In our Text-Guided Image Projection, we propose leveraging both referring text and visual tokens from each frame to guide the T2V model in producing the latent feature, rather than solely relying on visual tokens. This approach not only ensures visual feature consistency across time aiding temporal instance matching, but also enriches feature detail for better spatial differentiation. 
In Equation 1 presented in the paper, the referring text tokens employ the attention mechanism to guide the visual tokens to get final prompt. To demonstrate the effectiveness of our design, we contrast our approach with a method that concatenates referring text tokens and visual tokens as the prompt. It is noteworthy that after the concatenation operation, we follow up with an MLP to ensure the number of trainable parameters is consistent with those in the attention mechanism for a fair comparison. Table~\ref{tab:tim} reveals that mere concatenation results in the inability of U-Net to effectively utilize the details of visual tokens, hindering the extraction of fine-grained visual features.

\noindent\textbf{Noise Level.} 
As discussed in Section 4.1 of the paper, to extract visual features using the denoising U-Net, it is essential to add noise to the video signal. However, the optimal level of noise strength is unknown. Specifically, in the forward process of the diffusion model, the level of noise is represented by steps; the larger the step, the greater the intensity of the noise. Table~\ref{tab:nstep} reports that when the step is growing, the performance is dropping. We believe that higher noise levels might obscure the original signal, resulting in less detailed visual features, which in turn leads to poorer performance.

\end{document}